\documentclass[preprint,11pt,nopreprintline]{elsarticle}

\usepackage{amssymb}
\usepackage{amsmath}

\usepackage{newtxtext}
\usepackage{newtxmath}
\usepackage{geometry}
\usepackage{chngcntr}
\usepackage{booktabs}
\usepackage{pifont}
\usepackage{makecell}
\usepackage{adjustbox}
\usepackage[table]{xcolor}
\usepackage{wrapfig}
\usepackage{multirow}
\usepackage{collcell}
\newlength{\mycellwd}
\newcommand{\myrbox}[1]{\makebox[\mycellwd][r]{#1}}
\newcolumntype{Z}[1]{>{\setlength{\mycellwd}{#1}\collectcell\myrbox}c<{\endcollectcell}}

\usepackage{xspace}
\newcommand{\eg}{e.g.\@\xspace}
\newcommand{\ie}{i.e.\@\xspace}

\newcommand{\cf}{cf.\@\xspace}

\newcommand{\worst}[1]{\cellcolor{blue!25}{#1}}       % 最低
\newcommand{\secondworst}[1]{\cellcolor{blue!15}{#1}} % 第二低
\newcommand{\thirdworst}[1]{\cellcolor{blue!8}{#1}}   % 第三低

\newcommand{\pub}[1]{{\textcolor{gray}{\scriptsize[#1]}}}

\usepackage[colorlinks,linkcolor=blue,citecolor=blue,urlcolor=blue]{hyperref}
\usepackage[nameinlink,capitalise]{cleveref}

\newcommand{\revise}[1]{#1}

\journal{Pattern Recognition}

\begin{document}

\begin{frontmatter}

\title{Video Object Segmentation under Aquatic Domain Shift: A Diagnostic Benchmark and Restoration Adapter}
\nonumnote{This is the author accepted manuscript of an article published in \textit{Pattern Recognition}.
The version of record is available at \href{https://doi.org/10.1016/j.patcog.2026.115218}{doi:10.1016/j.patcog.2026.115218}.
\textcopyright{} 2026. This manuscript version is made available under the
\href{https://creativecommons.org/licenses/by-nc-nd/4.0/}{CC BY-NC-ND 4.0 license}.}

\author[aff1]{Hongshen Zhao} %% Author name
\author[aff1]{Jingkang Tai}
\author[aff1]{Yuhang Wu}
\author[aff2]{Wenkang Zhang}
\author[aff1]{Xi Lan}
\author[aff1]{Shangyan Wang}
\author[aff1]{Tianyu Zhang}
\author[aff1,aff3]{Wankou Yang}

\affiliation[aff1]{organization={School of Automation, Southeast University},%Department and Organization
               city={Nanjing},
               postcode={210096}, 
               country={China}}
\affiliation[aff2]{organization={School of Cyber Science and Engineering, Southeast University},
               city={Nanjing},
               postcode={210096},  
               country={China}}
\affiliation[aff3]{organization={Advanced Ocean Institute, Southeast University},
               city={Nantong},
               postcode={226010},  
               country={China}}

\begin{abstract}
Underwater video object segmentation (VOS) requires temporally consistent dense segmentation in scenes where water acts as a participating optical medium. Wavelength-dependent attenuation, scattering, turbidity, and unstable illumination alter object appearance, weaken mask-level localization cues, and destabilize temporal association, yet existing VOS benchmarks provide limited support for diagnosing these aquatic factors. We introduce \textbf{UW-VOS}, a large-scale diagnostic benchmark with 1,431 video sequences, 307,267 mask annotations, 411 semantic categories, and 16 fine-grained attributes. UW-VOS supports both standard VOS evaluation and factorized analysis of aquatic appearance degradation, spatial localization ambiguity, and temporal identity discontinuity. We further propose the \textbf{Aquatic Restoration Adapter} (ARA), a physics-derived parameter-efficient module that transfers the inverse structure of underwater image formation into bottleneck feature space and can be instantiated in different VOS backbones. Experiments show that representative open-air VOS methods suffer an average 13-point $\mathcal{J}\&\mathcal{F}$ drop on UW-VOS. ARA matches or surpasses full fine-tuning in both Transformer- and memory-based VOS backbones, and outperforms standard PEFT counterparts under a matched setting. These results establish UW-VOS as a diagnostic platform for underwater video perception and ARA as a compact feature-restoration module for aquatic domain adaptation.
Project repository: \url{https://github.com/hszh/UW-VOS}.
\end{abstract}


\end{frontmatter}

\section{Introduction}
\label{sec:intro}

\begin{figure*}[tb]
  \centering
  \includegraphics[width=\textwidth]{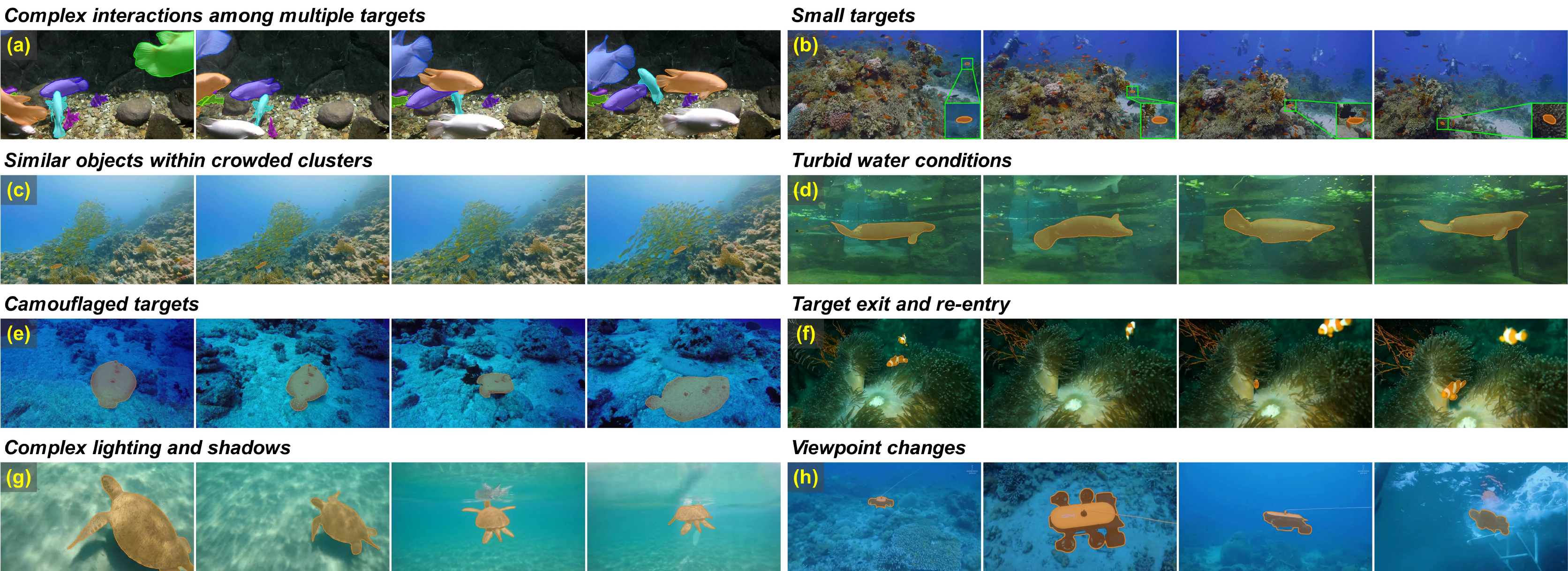}
  \caption{Example videos from the proposed UW-VOS dataset. The selected targets are highlighted by colored masks. The figure illustrates several representative characteristics in the UW-VOS dataset, including complex interactions among multiple targets (a), small objects within similar clusters (b, c), turbid water conditions (d), camouflaged targets (e), target exit and re-entry (f), complex lighting and shadows (g), as well as viewpoint changes (h).}
  \label{fig:ExampleVideos}
\end{figure*}

Underwater video object segmentation (VOS) studies temporal dense segmentation under aquatic domain shift, where the goal is to segment and track target instances throughout underwater videos given their initial masks. This task is important for marine biodiversity assessment~\cite{AquaOV255}, animal behavioral analysis~\cite{AlphaTracker}, and autonomous underwater vehicle navigation~\cite{an2026auv,kapoor2025underwater}. Beyond these applications, underwater VOS raises a central scientific question: how do aquatic image-formation effects propagate from degraded visual observations to dense mask localization and temporal identity preservation? Unlike terrestrial scenes, underwater videos are captured through a participating optical medium, where wavelength-dependent absorption, scattering, turbidity, and caustic illumination alter object appearance, boundary evidence, and feature correspondence. Consequently, underwater VOS is not merely open-air VOS applied to a new scene category, but a setting where image formation directly affects dense localization and temporal association.

We view underwater VOS failures as the interaction of three measurable factors. \textbf{Aquatic appearance degradation} changes the observed target appearance through wavelength-dependent attenuation, veiling light, illumination variation, low visibility, and water color bias. \textbf{Spatial localization ambiguity} weakens the mask-level evidence required for dense segmentation, especially when targets are small, motion-blurred, camouflaged, or surrounded by visually similar distractors. \textbf{Temporal identity discontinuity} breaks long-term association when organisms move unpredictably, become occluded, disappear, and re-enter the field of view. These factors are coupled: visibility degradation can amplify camouflage, small targets are more easily confused with similar instances, and occlusion often leads to identity switches after re-entry. A useful underwater VOS benchmark should therefore diagnose which factor causes failure, rather than only report an overall ranking.

Existing datasets and methods are insufficient for studying this problem systematically. General VOS methods~\cite{STM,STCN,XMem,Cutie,SAM2} have achieved strong performance on open-air benchmarks such as DAVIS~\cite{DAVIS16,DAVIS17}, YouTube-VOS~\cite{YouTube-VOS}, and MOSE~\cite{MOSEv1,MOSEv2}, but their evaluation protocols do not isolate aquatic appearance degradation, localization ambiguity, or temporal identity discontinuity. In the underwater video domain, research has primarily advanced along visual object tracking (VOT), progressing from UOT100~\cite{UOT100} and UTB180~\cite{UTB180} to the million-scale WebUOT-1M~\cite{WEBUOT1M}, while pixel-level video segmentation remains much less explored. Existing underwater VOS resources such as CoralVOS~\cite{CoralVOS} are restricted in category diversity and diagnostic attributes. As a result, it remains unclear which underwater factors dominate VOS failures, how different model families respond to them, and what form of domain adaptation is actually needed.

To enable such diagnosis, we introduce \textbf{UW-VOS}, a large-scale diagnostic benchmark for underwater video object segmentation. As illustrated in \cref{fig:ExampleVideos}, UW-VOS covers complex multi-target interactions, small objects within dense clusters, turbid water conditions, camouflaged targets, target exit and re-entry, complex lighting, and viewpoint changes. The dataset comprises 1,431 video sequences spanning 411 semantic categories with 307,267 mask annotations and 16 fine-grained attributes. Its annotations are produced through a human-in-the-loop pipeline with explicit quality auditing, and its attributes support factorized evaluation of appearance, localization, and temporal association failures. In this sense, UW-VOS is not only a training and evaluation resource, but also a diagnostic instrument for understanding aquatic domain shift in temporal dense segmentation.

Complementing this benchmark, we propose the \textbf{Aquatic Restoration Adapter} (ARA), a physics-derived parameter-efficient adapter for underwater VOS. Rather than treating underwater adaptation as an unconstrained feature tuning problem, ARA transfers the closed-form inverse of underwater image formation into bottleneck feature space. It estimates a content-dependent inverse-transmission gain and a channel-wise veiling-light term, yielding a constrained feature restoration operator that can be inserted into different VOS backbones. In this paper, we instantiate ARA on both SAM2~\cite{SAM2} and XMem~\cite{XMem} while keeping their temporal memory mechanisms unchanged. This controlled design allows us to ask how much underwater VOS performance can be improved by spatial feature restoration alone, and which remaining failures require temporal identity modeling beyond feature adaptation.

Experiments on UW-VOS provide a factorized view of underwater VOS difficulty. Representative open-air VOS methods suffer a substantial performance drop when transferred to underwater videos, confirming the need for a dedicated diagnostic benchmark. In-domain experiments show that ARA matches or surpasses full fine-tuning on both SAM2~\cite{SAM2} and XMem~\cite{XMem}, indicating that the proposed adapter is not tied to a specific backbone family. Attribute-level analysis reveals that ARA mainly improves aquatic appearance degradation and spatial localization ambiguity, while temporal identity discontinuity remains a complementary challenge for future underwater VOS research.

In summary, our main contributions are as follows:
\begin{itemize}
\setlength\itemsep{0.2em}
    \item We formulate underwater VOS as temporal dense segmentation under aquatic domain shift, decomposing its failures into aquatic appearance degradation, spatial localization ambiguity, and temporal identity discontinuity.

    \item We present UW-VOS, a large-scale diagnostic benchmark with 1,431 videos, 411 categories, 307,267 mask annotations, and 16 fine-grained attributes, together with quality-audited annotations and factorized evaluation protocols.

    \item We propose ARA, a physics-derived parameter-efficient adapter that implements an underwater image-formation inverse in bottleneck feature space and can be instantiated in both Transformer and CNN-based VOS backbones.

    \item We establish a comprehensive evaluation suite covering cross-domain transfer, in-domain training, full fine-tuning and PEFT comparisons, response visualization, and attribute-based diagnosis, quantifying the underwater domain gap and validating ARA as a parameter-efficient feature-space restoration baseline.
\end{itemize}

\section{Related Work}
\subsection{Video Object Segmentation}

Semi-supervised video object segmentation aims to segment and track target instances throughout a video given their first-frame masks. Early methods relied on online adaptation or recurrent propagation, which often suffered from slow inference or error accumulation~\cite{RecurrentVOS,swiftnet,OSVOS,PReMVOS}. Space-Time Memory networks established a matching-based framework that stores past frames as memory~\cite{STM}, and later methods refined this paradigm through improved memory management~\cite{STCN,XMem,CFBI,CFBI+}, hierarchical propagation transformers~\cite{AOT,DeAOT}, object-level reasoning~\cite{Cutie}, and unified spatio-temporal modeling~\cite{SimVOS,VITA,JointFormer}. More recently, SeC~\cite{SeC} introduced progressive concept construction for complex appearance changes and scene transitions, while SAM2~\cite{SAM2} extended promptable segmentation to videos through a streaming memory architecture. These methods provide strong temporal association and mask propagation mechanisms, but they are primarily developed and evaluated on open-air videos. Consequently, their benchmarks and adaptation strategies do not explicitly diagnose how underwater image formation affects appearance, localization, and identity continuity. Our work addresses this gap by evaluating representative VOS methods under aquatic domain shift and by introducing ARA as a feature-space restoration adapter that can be instantiated without redesigning the host temporal memory.

\subsection{Underwater Visual Perception}

Underwater visual perception has evolved from image-level recognition to instance-level and video-level understanding. Early benchmarks such as DeepFish~\cite{DeepFish} focused on classification, whereas recent datasets cover underwater instance and salient object segmentation~\cite{UIIS,UIIS10K,USIS10K,USIS16K}, camouflaged instance segmentation~\cite{UCIS4K}, open-set recognition~\cite{AquaOV255,MARIS}, and multi-task instruction tuning~\cite{NautData}. In the video domain, most underwater resources have been developed for visual object tracking, progressing from UOT100~\cite{UOT100} to attribute-rich benchmarks such as UTB180~\cite{UTB180}, UVOT400~\cite{UVOT400}, and UWCOT220~\cite{UWCOT220}, and then to large-scale or multi-modal resources such as WebUOT-1M~\cite{WEBUOT1M} and MSC~\cite{MSC}. These datasets substantially advance underwater perception, but their supervision is typically image-level, box-level, language-grounded, or restricted to specific domains. Pixel-level underwater VOS remains comparatively underdeveloped: CoralVOS~\cite{CoralVOS} provides an important step toward underwater mask propagation, but its category diversity and diagnostic coverage are limited. UW-VOS is designed to fill this missing benchmark role by providing large-scale mask annotations, diverse categories, and fine-grained attributes for analyzing aquatic appearance degradation, spatial localization ambiguity, and temporal identity discontinuity.

\section{UW-VOS Dataset}
\subsection{Video Collection and Annotation}
To construct a highly diverse and precisely annotated benchmark for underwater video object segmentation, we designed a semi-automatic data engine that integrates image restoration with foundation model assistance, as illustrated in \cref{fig:DataEngine}. This pipeline comprises three stages:

\begin{figure}[tb]
  \centering
  \includegraphics[height=3.5cm]{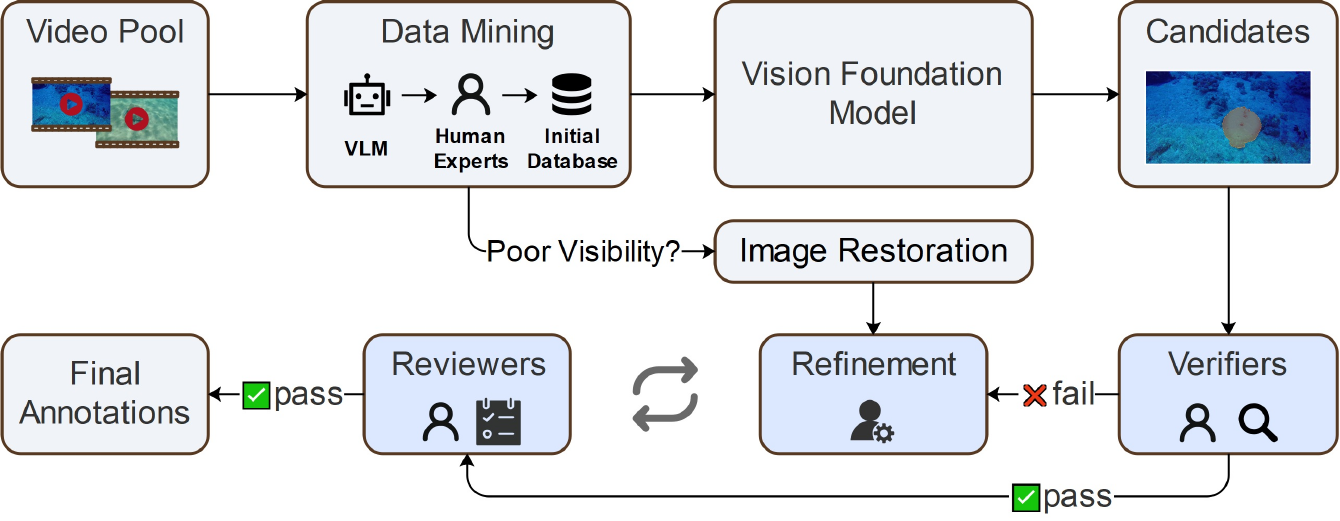}
  \caption{Overview of the semi-automatic data engine designed for constructing the UW-VOS dataset. The final annotations have undergone multiple rounds of manual verification (highlighted in light blue).}
  \label{fig:DataEngine}
\end{figure}

\textbf{Video Acquisition and Data Mining.} UW-VOS comprises 1,032 videos derived from WebUOT-1M~\cite{WEBUOT1M} and 399 newly mined copyright-free Internet videos. The source split concerns video acquisition only: the WebUOT-1M-derived subset provides tracking-style annotations but no pixel-level VOS masks in our target format. After source-level and visual-content deduplication, all 307,267 final pixel-level masks and associated attribute annotations are newly produced under our unified annotation pipeline; detailed provenance statistics are provided in the Appendix. During data mining, we implement a difficulty-oriented active selection strategy that prioritizes challenging scenarios involving rare species, frequent viewpoint changes, occlusions, small targets, similar distractors, and biological camouflage, while excluding simple samples with monotonous target motion or overly homogeneous backgrounds. We then identify targets of interest within each video and provide first-frame masks, with semantic categories determined by combining vision-language models~\cite{gemini2.5} and human expert knowledge to ensure rich biological and non-biological diversity.

\textbf{Foundation Model-Assisted Annotation.} Using the established first-frame masks, we leverage advanced visual foundation models~\cite{SAM2} to automatically perform temporal propagation and generate candidate masks across all sequences. Given the color distortion and turbidity inherent to underwater scenes, we integrate an image restoration module~\cite{SemiUIR} only as an auxiliary visual reference for ambiguous or low-confidence samples. Restored frames are shown side-by-side with the original frames during human verification, but the final masks are checked and stored in the original frame coordinates. Restored frames are not used for model training or benchmark evaluation, which avoids introducing enhancement artifacts into the released ground truth or experimental protocol.

\textbf{Human-in-the-Loop Verification and Refinement.} Human verification is performed by an eight-member annotation team. Each video sequence is assigned to one annotator, who checks the candidate attribute labels obtained during data mining and the masks propagated by the visual foundation model. Masks exhibiting identity switches, missing targets, false positives, or boundary drift are interactively corrected on the original frames and rechecked by the assigned annotator. Cases with ambiguous attribute labels or uncertain mask quality are assigned to one of two domain experts for final adjudication.

\textbf{Annotation Quality Audit.} Separately from the production workflow, we conduct a stratified reliability audit on 100 complete video sequences (20\,K frames) spanning data sources, categories, object scales, and challenge attributes. Two annotators independently verify the same audit set. Their masks yield an average inter-annotator $\mathcal{J}\&\mathcal{F}$ of 96.3. For attribute labels, we compute Cohen's $\kappa$~\cite{Cohen1960} separately for each of the 16 attributes and report the macro-averaged value of 0.82. Relative to the foundation-model candidates, 19\% of audited masks are accepted unchanged, 48\% require local boundary refinement, and 33\% require reconstruction due to substantial omissions, false positives, or drift; 4\% of audited sequences additionally require identity correction. The high agreement supports annotation consistency, while the correction rates underscore the need for human verification beyond foundation-model propagation.

\subsection{Dataset Statistics}

\begin{table}[tb]
  \centering
  \caption{Comparison of VOS datasets. $\ast$ indicates the segmentation subset.}
  \label{tab:datasets-comparison}
  \begin{adjustbox}{max width=\textwidth}
    \begin{tabular}{
        l@{\hspace{5pt}}
        c@{\hspace{15pt}}  % Year
        Z{2.6em}@{\hspace{12pt}}  % Videos
        Z{2.6em}@{\hspace{10pt}}  % Categ
        Z{2.6em}@{\hspace{10pt}}  % Attr
        Z{2.6em}@{\hspace{10pt}}  % Mean
        Z{3.0em}@{\hspace{12pt}}  % Max
        Z{3.5em}@{\hspace{10pt}}  % Total
        Z{3.0em}  % Annot
    }
      \toprule
      Dataset & 
      Year & 
      \multicolumn{1}{l}{Videos} & 
      \multicolumn{1}{l}{Classes} & 
      \multicolumn{1}{l}{Attributes} & 
      \multicolumn{1}{l}{\makecell{Mean\\frame}} & 
      \multicolumn{1}{l}{\makecell{Max\\frame}} & 
      \multicolumn{1}{l}{\makecell{Total\\frames}} & 
      \multicolumn{1}{l}{Annotations} \\
      \midrule
      \multicolumn{9}{c}{{\textit{General Scenarios}}} \\
      \midrule
      DAVIS$_{16}$~\cite{DAVIS16} & 2016 & 50 & - & 15 & 69 & 99 & 3\,K & 3\,K  \\
      DAVIS$_{17}$~\cite{DAVIS17} & 2017 & 90 & - & 15 & 69 & 104 & 6\,K & 13\,K  \\
      YouTube-VOS~\cite{YouTube-VOS} & 2018 & 4,453 & 94 & 0 & 27 & - & 120\,K & 197\,K  \\
      VOST~\cite{VOST} & 2023 & 713 & 155 & 51 & 106 & - & 75\,K & 173\,K  \\
      LVOS~\cite{LVOS} & 2023 & 220 & 27 & 13 & 574 & - & 126\,K & 156\,K  \\
      MOSEv1~\cite{MOSEv1} & 2023 & 2,149 & 36 & 0 & 61 & - & 130\,K & 431\,K  \\
      MOSEv2~\cite{MOSEv2} & 2025 & 5,024 & 200 & 15 & 93 & 7,825 & 468\,K & 701\,K  \\
      \midrule
      \multicolumn{9}{c}{{\textit{Underwater Scenarios}}} \\
      \midrule
      DeepFish$^\ast$~\cite{DeepFish} & 2020 & 19 & 1 & 0 & 33 & 79 & 0.6\,K & 0.3\,K  \\
      Seagrass~\cite{Seagrass} & 2021 & 138 & 2 & 0 & 31 & 278 & 4\,K & 9\,K  \\
      CoralVOS~\cite{CoralVOS} & 2025 & 150 & 1 & 0 & 403 & - & 60\,K & 60\,K  \\
      \midrule
      \textbf{UW-VOS} & \textbf{2026} & \textbf{1,431} & \textbf{411} & \textbf{16} & \textbf{201} & \textbf{3,329} & \textbf{287,199} & \textbf{307,267}  \\
      \bottomrule
    \end{tabular}
  \end{adjustbox}
\end{table}

In \cref{tab:datasets-comparison}, we compare UW-VOS with existing VOS datasets across general and underwater scenarios. VOS datasets for general scenes have witnessed significant growth in recent years, both in category diversity and annotation volume, exemplified by YouTube-VOS~\cite{YouTube-VOS} and MOSEv2~\cite{MOSEv2}. In contrast, existing underwater VOS datasets remain severely limited: DeepFish~\cite{DeepFish} contains only 19 videos of a single category, Seagrass~\cite{Seagrass} covers merely 2 categories across 138 videos, and CoralVOS~\cite{CoralVOS} is restricted to a single category despite relatively longer sequences; notably, none of them provides attribute annotations for fine-grained diagnostic evaluation. UW-VOS addresses these limitations by providing 1,431 video sequences spanning 411 semantic categories with 307,267 mask annotations and 16 fine-grained attributes for systematic performance analysis, achieving order-of-magnitude improvements in video quantity, category diversity, and annotation density, and establishing a comprehensive benchmark for underwater video object segmentation.

\begin{figure}[tb]
  \centering
  \includegraphics[width=\textwidth]{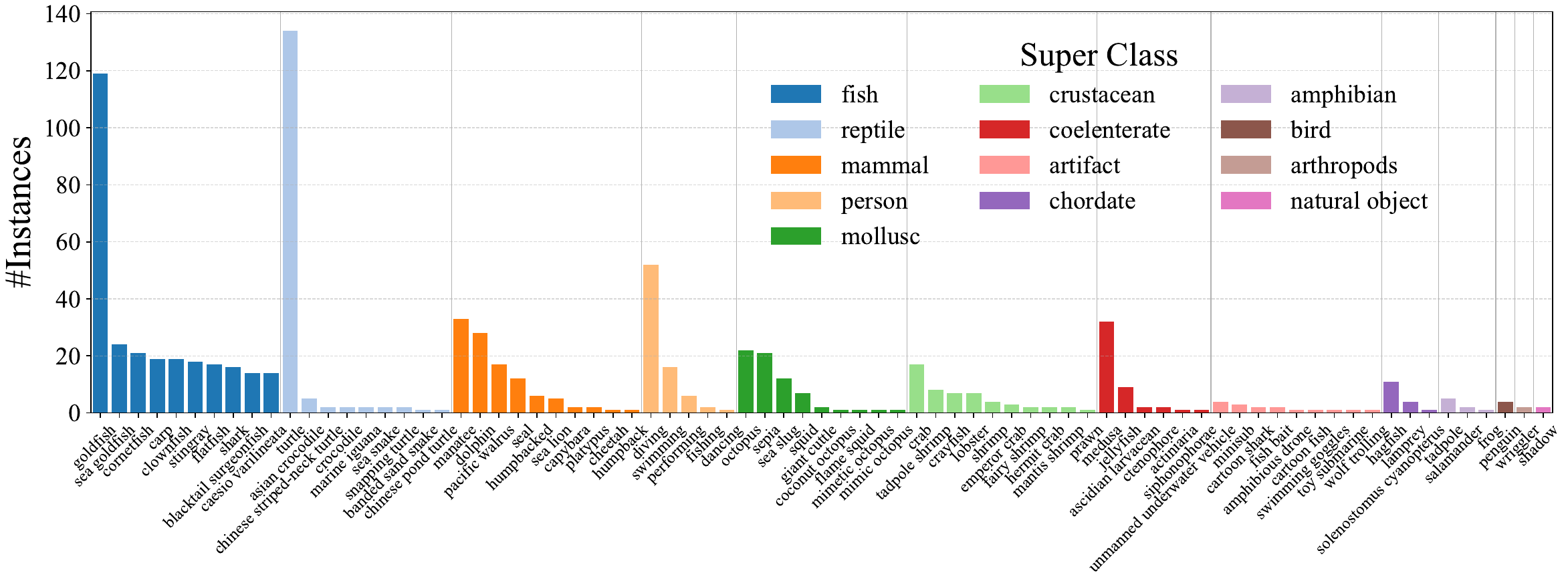}
  \caption{Category-instance distribution of the UW-VOS dataset. Due to space constraints, we only visualize the top 10 subcategories within each superclass.}
  \label{fig:ClassInstance}
\end{figure}

\textbf{Categories.} UW-VOS includes 411 semantic categories grouped into 13 superclasses, covering a wide range of biological entities and non-biological objects. As shown in \cref{fig:ClassInstance}, these categories span fish, reptiles, mammals, persons, molluscs, crustaceans, coelenterates, chordates, amphibians, birds, arthropods, artifacts, and natural objects. The category distribution is highly long-tailed: common classes such as goldfish contain more than 120 instances, whereas many rare species have only a few samples. This imbalance makes robust generalization challenging for both frequent and rare categories.

\begin{figure*}[tb] \centering
\begin{minipage}[t]{0.59\linewidth}
    \includegraphics[width=\textwidth]{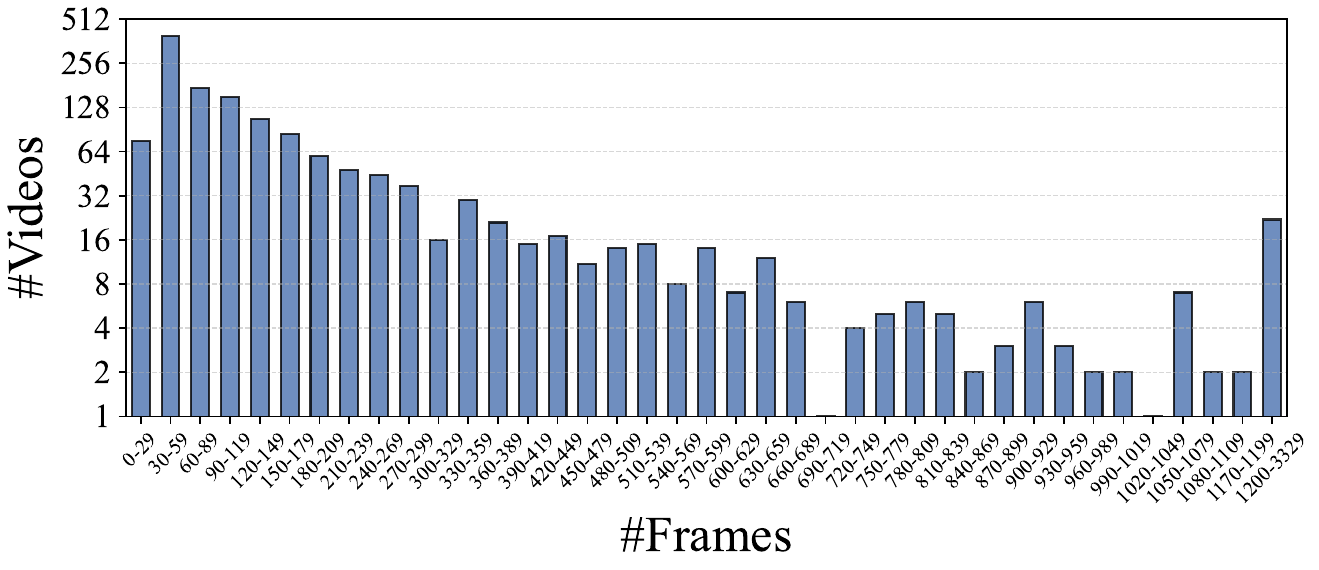}
    \caption{Video length distribution across diverse frame-count ranges.} \label{fig:video_length}
\end{minipage}\hfill
\begin{minipage}[t]{0.38\linewidth}
    \includegraphics[width=\textwidth]{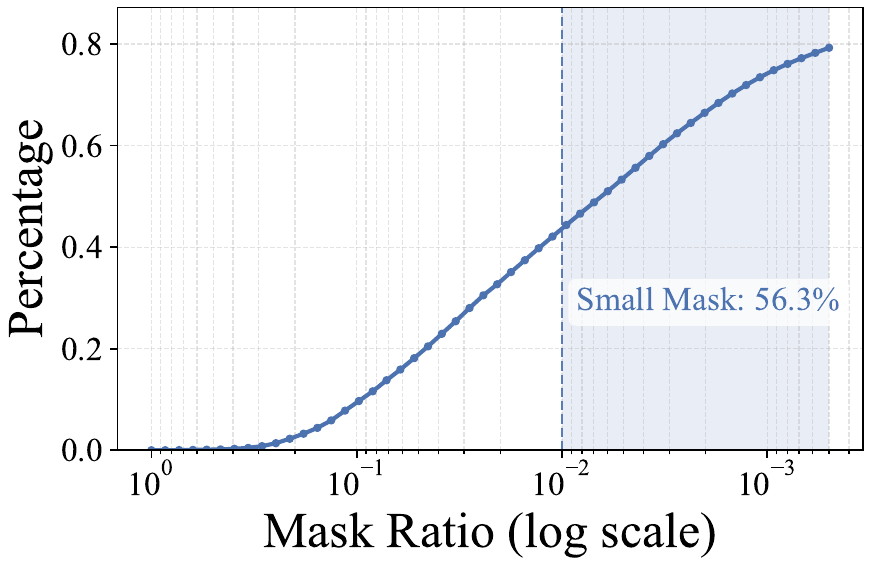}
    \caption{Mask size distribution, normalized by video resolution.} \label{fig:mask_size}
\end{minipage}
\end{figure*}

\textbf{Video Length.} \Cref{fig:video_length} shows the video length distribution of UW-VOS across diverse frame-count ranges, spanning from short clips to extended sequences with over 1,200 frames. The dataset contains 1,431 video sequences with an average of 201 frames per video and a maximum sequence length of 3,329 frames, maintaining substantial coverage across all frame-count intervals. This broad distribution of video lengths facilitates a rigorous assessment of temporal consistency and long-term tracking robustness, which is especially crucial in underwater environments, where prolonged tracking is severely challenged by continuous perturbations such as water turbidity, illumination shifts, and frequent occlusions and re-appearances.

\textbf{Mask Size.} \Cref{fig:mask_size} illustrates the mask size distribution in UW-VOS. Following \cite{MOSEv2}, we normalize mask areas relative to video resolution to account for varying resolutions. Small masks (\ie, mask ratio $<$ 0.01) constitute 56.3\% of all instances, significantly exceeding public datasets such as MOSEv2 (50.2\%), DAVIS (25.3\%), and YouTube-VOS (18.4\%). This prevalence of small objects reflects a critical challenge in underwater video segmentation, emphasizing the necessity for robust detection and tracking of small targets within complex underwater environments.

\begin{figure*}[tb] \centering
\begin{minipage}[t]{0.35\linewidth}
    \includegraphics[width=\textwidth]{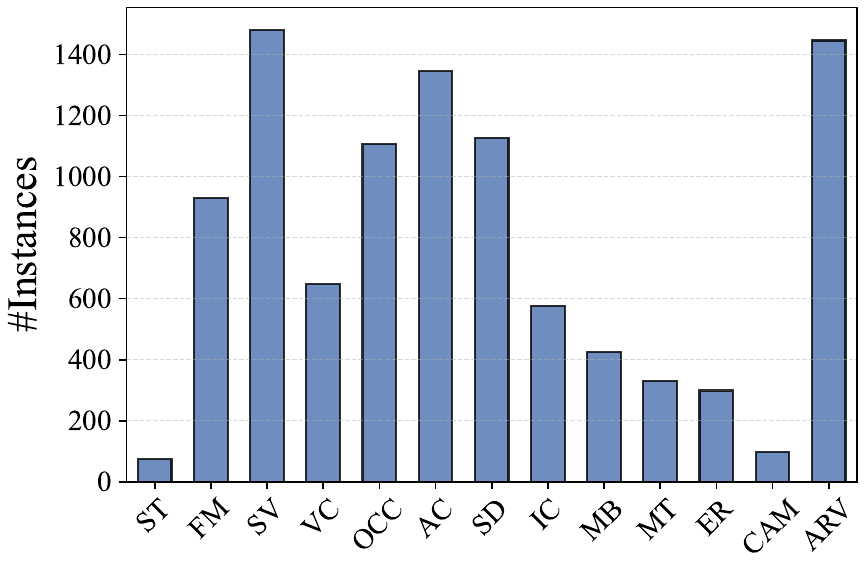}
    \caption{Distribution of instance counts across attributes.} \label{fig:attribute_instance}
\end{minipage}\hfill
\begin{minipage}[t]{0.3\linewidth}
    \includegraphics[width=\textwidth]{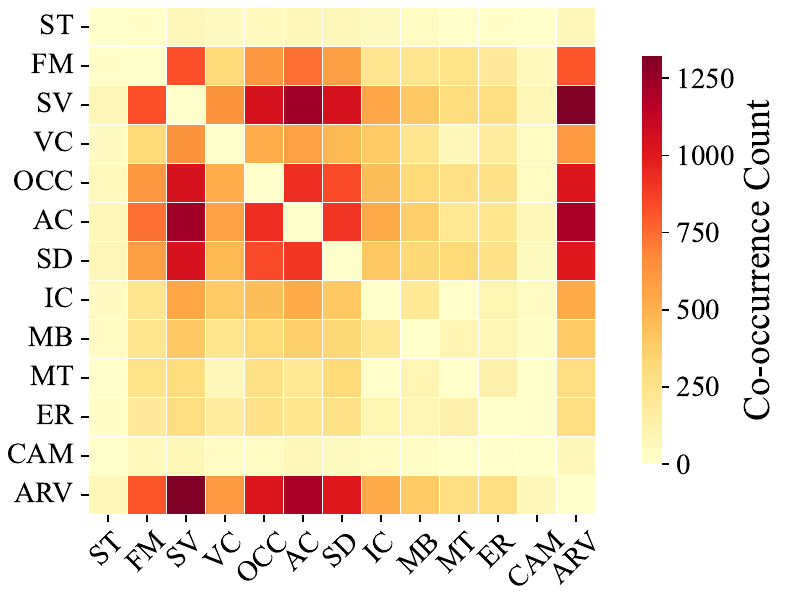}
    \caption{Co-occurrence count heatmap of binary attributes.} \label{fig:attribute_heatmap}
\end{minipage}\hfill
\begin{minipage}[t]{0.3\linewidth}
    \includegraphics[width=\textwidth]{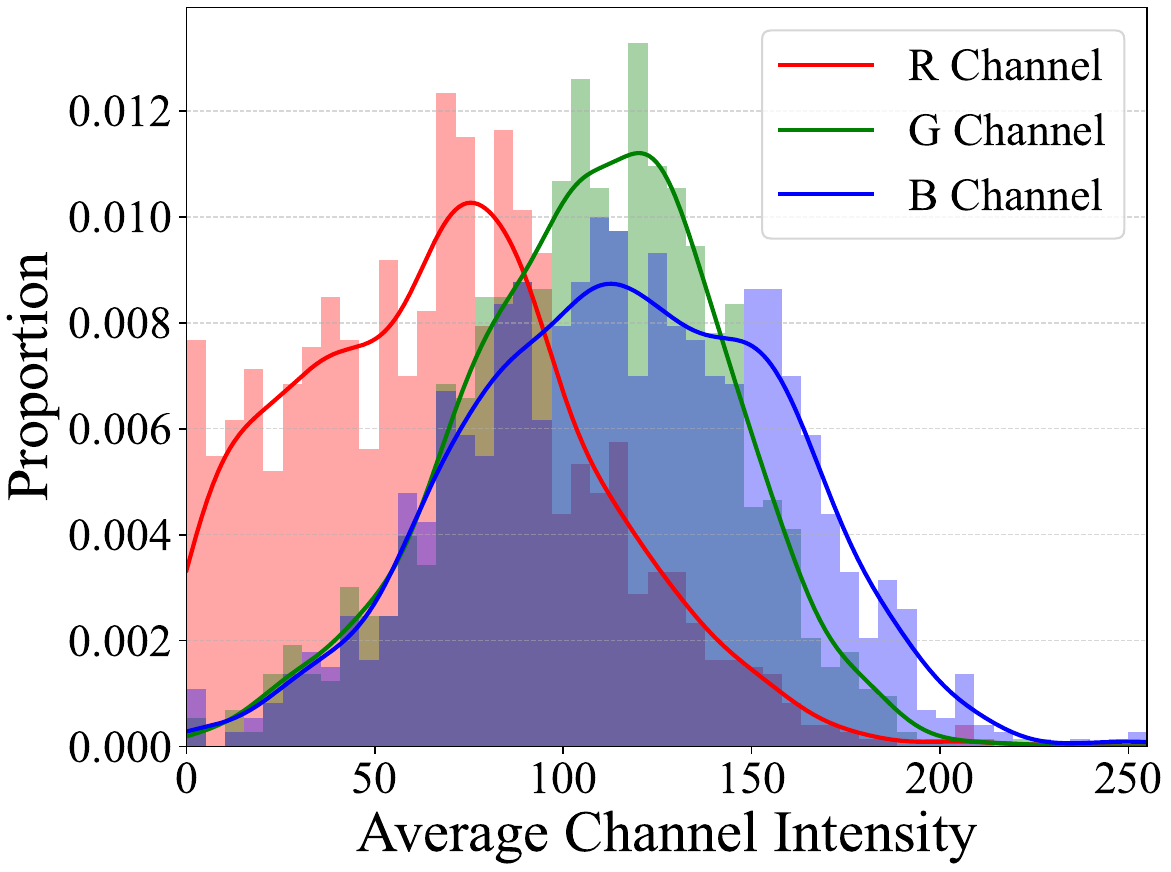}
    \caption{Distribution of average RGB channel intensities.}
    \label{fig:channel_intensity}
\end{minipage}
\end{figure*}

\textbf{Attribute Analysis.} We annotate each object instance with 16 attributes to enable comprehensive performance analysis under diverse challenging conditions (see the Appendices for detailed definitions). \Cref{fig:attribute_instance} presents the distribution of instance counts across 13 binary attributes. The high prevalence of scale variation (SV), aspect ratio variation (ARV), appearance change (AC), occlusion (OCC), similar distractors (SD), and fast motion (FM) collectively reflects the intrinsic complexity of underwater environments, where aquatic organisms exhibit rapid movements, continuous shape deformations, and frequent interactions with visually similar species. Notably, small target (ST) and camouflage (CAM), despite lower occurrence, represent critical challenges unique to underwater scenarios where tiny objects and background similarity significantly impede segmentation accuracy. \Cref{fig:attribute_heatmap} further visualizes the co-occurrence patterns among these attributes. Strong co-occurrence relationships are observed between SV-AC, SV-ARV, and FM-SD, indicating that underwater objects often undergo simultaneous geometric and appearance transformations. The underwater-specific attribute ARV shows high co-occurrence with multiple generic attributes, highlighting the compound challenges in underwater video object segmentation. 

\textbf{Underwater Color Characteristics.} \Cref{fig:channel_intensity} depicts the distribution of average RGB channel intensities in the first frame of all videos. The green channel exhibits the highest intensity distribution with peak around 120, while the red channel shows significant attenuation with peak around 80, consistent with wavelength-dependent light absorption in water where red light attenuates most rapidly. This pronounced color shift toward blue-green spectrum diminishes inter-object color contrast and exacerbates the camouflage and similar distractor challenges discussed above, posing additional difficulties for appearance-based segmentation methods.

\section{Methodology}

\subsection{Overview}
\begin{figure}[tb]
  \centering
  \includegraphics[width=\textwidth]{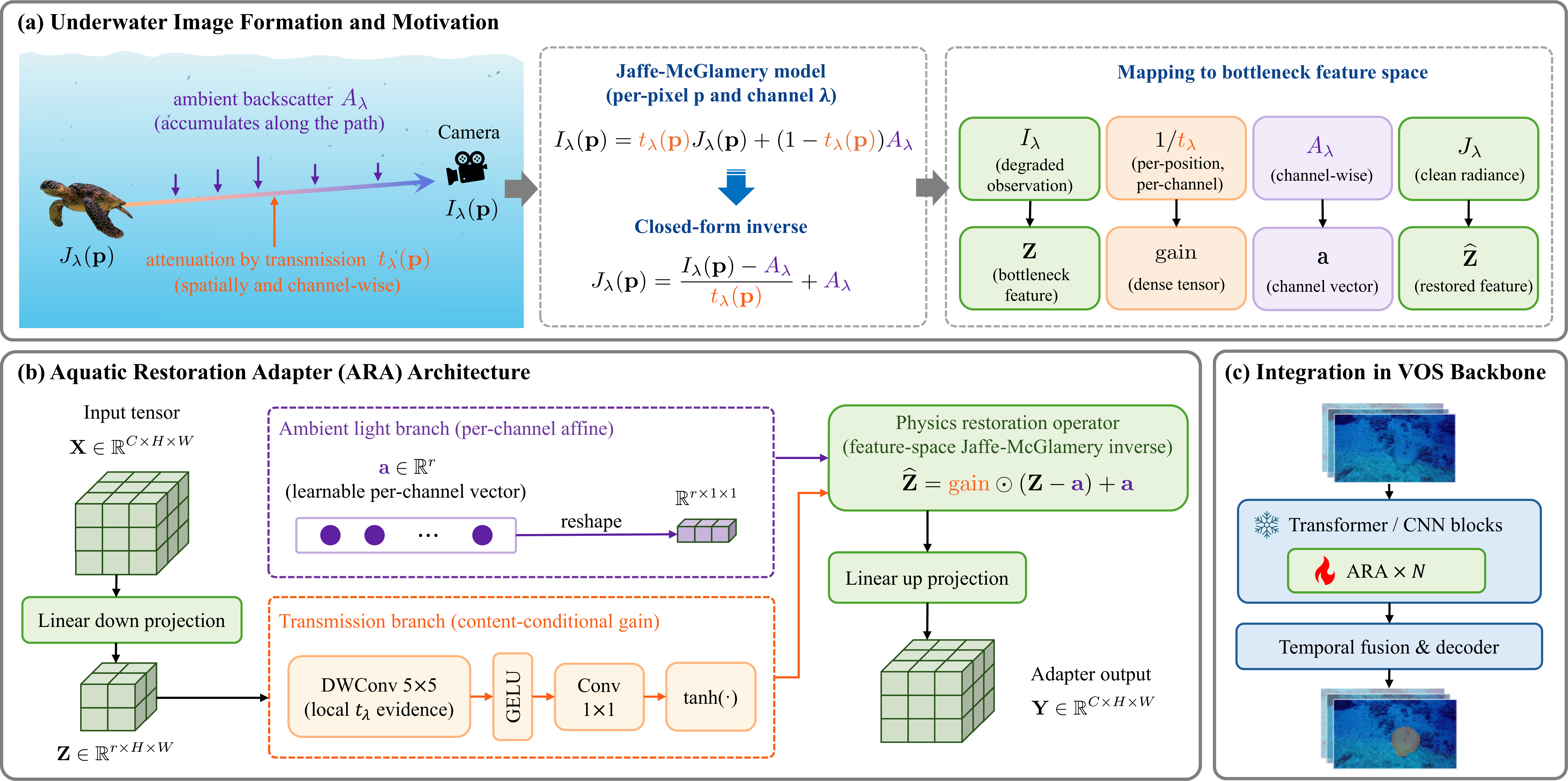}
  \caption{Overview of the proposed Aquatic Restoration Adapter (ARA). (a) Underwater image formation motivates a feature-space inverse restoration operator. (b) ARA estimates a content-dependent inverse transmission gain and a channel-wise veiling-light term in the bottleneck feature space. (c) ARA can be inserted into different VOS backbones.}
  \label{fig:ara}
\end{figure}

We propose the \textbf{Aquatic Restoration Adapter} (ARA), a physics-derived parameter-efficient module for underwater feature adaptation, as shown in \Cref{fig:ara}. ARA is not tied to a particular VOS architecture: given a spatial feature map from either a Transformer or CNN backbone, it applies a constrained restoration operation in a low-dimensional bottleneck space and returns a residual update to the original feature stream. The key idea is to transfer the algebraic structure of underwater image formation inversion from the image domain to the feature domain, so that the adapter corrects water-induced feature degradation without modifying the input frames or fully fine-tuning the backbone. In this work, we instantiate ARA in two representative VOS backbones, SAM2~\cite{SAM2} and XMem~\cite{XMem}, while keeping their temporal memory mechanisms unchanged. This design isolates how much underwater VOS performance can be improved by ARA-based spatial feature adaptation alone.

\subsection{From Underwater Image Formation to Feature Restoration}

ARA is motivated by the Jaffe--McGlamery underwater image formation model~\cite{McGlamery1980,Jaffe2002}. For a pixel location $\mathbf{p}$ and wavelength $\lambda$, a simplified formation process can be written as
\begin{equation}
  I_{\lambda}(\mathbf{p})
  =
  t_{\lambda}(\mathbf{p})J_{\lambda}(\mathbf{p})
  +
  \bigl(1-t_{\lambda}(\mathbf{p})\bigr)A_{\lambda},
  \quad
  t_{\lambda}(\mathbf{p})
  =
  \exp\!\bigl(-\eta_{\lambda}d(\mathbf{p})\bigr),
  \label{eq:underwater-formation}
\end{equation}
where $I_{\lambda}(\mathbf{p})$ is the observed underwater irradiance, $J_{\lambda}(\mathbf{p})$ is the latent scene radiance before water degradation, $t_{\lambda}(\mathbf{p})$ is the transmission, $A_{\lambda}$ denotes the global ambient or veiling light, $\eta_{\lambda}$ is the wavelength-dependent attenuation coefficient, and $d(\mathbf{p})$ is scene depth. This model captures two properties that are especially relevant to VOS. First, transmission is spatially varying because depth and viewing geometry change across the scene. Second, attenuation and veiling light are channel dependent because water absorbs different wavelengths at different rates. Solving \cref{eq:underwater-formation} for the latent radiance gives
\begin{equation}
  J_{\lambda}(\mathbf{p})
  =
  \frac{I_{\lambda}(\mathbf{p})-A_{\lambda}}
       {t_{\lambda}(\mathbf{p})}
  +
  A_{\lambda}.
  \label{eq:underwater-inverse}
\end{equation}
ARA implements the structure of \cref{eq:underwater-inverse} in feature space. We do not assume that a feature channel corresponds to a physical wavelength, nor do we perform image-level restoration. Instead, we use the inverse formation model as an inductive bias: subtract a channel-dependent veiling component, apply a spatially varying multiplicative gain that approximates inverse transmission, and restore the veiling level. This yields a physically constrained residual adapter rather than an unconstrained bottleneck transformation.

\subsection{Aquatic Restoration Adapter}

Given a spatial feature map $\mathbf{X}\in\mathbb{R}^{C\times H\times W}$, ARA first projects it into a bottleneck space:
\begin{equation}
  \mathbf{Z}=P_{\mathrm{down}}(\mathbf{X}), \quad
  \mathbf{Z}\in\mathbb{R}^{r\times H\times W},
  \label{eq:ara-down}
\end{equation}
where $r\ll C$ controls the trainable parameter budget. The adapter then estimates local evidence for transmission with a depth-wise spatial aggregator,
\begin{equation}
  \mathbf{C}_{T}
  =
  \sigma\!\left(\mathrm{DWConv}_{k\times k}(\mathbf{Z})\right),
  \label{eq:ara-context}
\end{equation}
where $\sigma$ denotes GELU. The use of local aggregation is important because transmission cannot be inferred reliably from an isolated pixel or token: the observed signal entangles latent appearance, depth, and veiling light. A depth-wise convolution provides local context while preserving channel-wise independence, matching the channel-asymmetric nature of underwater attenuation.

From this local context, ARA predicts a content-dependent inverse-transmission gain:
\begin{equation}
  \mathbf{G}
  =
  1+\tanh\!\left(P_{g}(\mathbf{C}_{T})\right),
  \quad
  \mathbf{G}\in\mathbb{R}^{r\times H\times W},
  \label{eq:ara-gain}
\end{equation}
where $P_g$ is a $1\times1$ projection. The gain is spatially varying and channel specific, reflecting that underwater transmission changes with depth, viewing geometry, and wavelength-dependent attenuation. In contrast, the veiling-light term is parameterized as a channel-wise vector $\mathbf{a}\in\mathbb{R}^{r}$ shared across spatial locations, following the standard assumption that ambient light is approximately global within a local scene. The bottleneck feature is restored by
\begin{equation}
  \widehat{\mathbf{Z}}
  =
  \mathbf{G}\odot
  \bigl(\mathbf{Z}-\mathbf{a}\bigr)
  +
  \mathbf{a},
  \quad
  \mathrm{ARA}(\mathbf{X})
  =
  P_{\mathrm{up}}(\widehat{\mathbf{Z}}),
  \label{eq:ara-restore}
\end{equation}
where $\odot$ denotes element-wise multiplication with broadcasting over spatial positions.

Intuitively, $\mathbf{G}$ performs content-dependent recalibration of the bottleneck responses according to local evidence of water-induced degradation, whereas $\mathbf{a}$ provides a spatially shared channel-wise reference for scene-level veiling effects. Their combination centers $\mathbf{Z}$ around this reference, applies a spatially varying adjustment, and then restores the reference level, thereby translating the inverse-formation structure into a feature-space operation. Accordingly, $\mathbf{G}$ and $\mathbf{a}$ should be understood as feature-space analogues motivated by the physical model rather than calibrated estimates of transmission and ambient light.

\revise{ARA uses the standard down/up projection scaffold of a bottleneck adapter; its novelty lies in the physically motivated, constrained feature transformation between these projections.} A standard branch produces the residual $P_{\mathrm{up}}(\phi(\mathbf{Z}))$ through a point-wise nonlinear bottleneck mapping. ARA instead predicts $\mathbf{G}$ from depth-wise local context and forms $\widehat{\mathbf{Z}}=\mathbf{G}\odot\mathbf{Z}+(1-\mathbf{G})\odot\mathbf{a}$ before $P_{\mathrm{up}}$, where $\mathbf{a}$ is a spatially shared channel vector. Consequently, the additive component is tied to the same content-dependent gain that scales $\mathbf{Z}$ rather than being generated independently. This coupling of local spatial aggregation, a spatially varying gain, and a constrained offset constitutes ARA's structural and algebraic distinction from a generic bottleneck adapter.

We initialize ARA to preserve the pretrained host at the start of fine-tuning. All weights and biases of $P_{\mathrm{up}}$ and $P_g$, as well as every entry of $\mathbf{a}$, are initialized exactly to zero, \ie, $\mathbf{a}=\mathbf{0}\in\mathbb{R}^{r}$ rather than a small constant. Thus, $\mathbf{G}=\mathbf{1}$ and the ARA residual initially vanishes because $P_{\mathrm{up}}=0$. In contrast, the depth-wise aggregator uses a delta kernel (unit center and zeros elsewhere) with zero bias, yielding $\mathbf{C}_{T}=\sigma(\mathbf{Z})$ and avoiding the permanent gradient blocking that a zero-initialized aggregator would cause in the content-conditioned gain pathway.

\subsection{Instantiation in VOS Backbones}

ARA can be inserted into any VOS backbone that exposes spatial feature maps. For a Transformer backbone, we instantiate ARA in the Hiera image encoder of SAM2~\cite{SAM2,Hiera}. The encoder is organized into four stages with progressively increasing channel dimensions. We keep the early stages frozen and insert ARA into Stage~3 and Stage~4, where object-level structure and domain-sensitive semantics are more strongly encoded. Within each selected Transformer block, one ARA branch is attached to the attention residual stream and another branch is attached to the feed-forward stream. All other SAM2 components, including the prompt encoder, mask decoder, and streaming temporal memory, remain frozen. This yields a parameter-efficient SAM2 instantiation with only a small fraction of trainable parameters while preserving the original video segmentation interface.

For a CNN and memory-based backbone, we instantiate ARA in XMem~\cite{XMem}. ARA is attached to the spatial feature stream before memory key-value construction, so that the features written to and queried from memory are corrected for water-induced degradation. The memory read, memory update, and temporal propagation modules are left unchanged. This XMem instantiation tests whether the same physics-constrained feature restoration operator transfers beyond Transformer foundation models. Across both SAM2 and XMem, ARA adapts only the spatial visual representation and deliberately leaves temporal association mechanisms intact. As a result, improvements can be attributed to feature-space aquatic restoration, while remaining failures on occlusion or exit-re-entry can still be analyzed as temporal identity discontinuity by UW-VOS.

\section{Experiments}
\subsection{Implementation Details}

The proposed UW-VOS dataset adopts the same data format as YouTube-VOS~\cite{YouTube-VOS}, facilitating seamless integration with existing VOS frameworks. Following standard practice, we partition the dataset into 1,001 training videos, 215 validation videos, and 215 test videos for model training, daily evaluation, and competition-period assessment, respectively.

\textbf{Evaluation Protocol.} Following the DAVIS protocol~\cite{DAVIS16,DAVIS17}, we evaluate model performance on the UW-VOS validation set using standard metrics: region similarity $\mathcal{J}$, contour accuracy $\mathcal{F}$, and their mean $\mathcal{J}\&\mathcal{F}$. To better capture the complex challenges inherent in UW-VOS, we additionally report $\dot{\mathcal{F}}$ and $\mathcal{J}\&\dot{\mathcal{F}}$ proposed in MOSEv2~\cite{MOSEv2}.

\textbf{Experimental Setup.} We conduct experiments under two settings to comprehensively evaluate model performance. For zero-shot cross-domain experiments, we uniformly employ model weights pretrained on open-air domain datasets, specifically DAVIS 2017~\cite{DAVIS17} and YouTube-VOS 2019~\cite{YouTube-VOS}, without any fine-tuning on underwater data. This setting evaluates the generalization capability of existing VOS methods when directly applied to the underwater domain. For in-domain experiments, all methods are trained using image-pretrained backbone without leveraging any additional video datasets. To ensure fair comparison, we follow the training protocol of MOSEv2~\cite{MOSEv2}, replacing the YouTube-VOS training set with our UW-VOS training set while strictly adhering to the original YouTube-VOS training configuration, including optimizer settings, learning rate schedules, and data augmentation strategies.

\subsection{Benchmark Results}

\begin{table*}[tb]
\caption{
Cross-domain evaluation results on UW-VOS validation set. We report the mean $\mathcal{J}\&\mathcal{F}$ on DAVIS$_{17}$ and YouTube-VOS$_{19}$ as open-air reference; values in parentheses indicate the gap relative to UW-VOS $\mathcal{J}\&\mathcal{F}$.
}
\centering
{\fontsize{10pt}{3.6mm}\selectfont
\setlength{\tabcolsep}{1.3mm}

\begin{adjustbox}{max width=\textwidth}
\begin{tabular}{l|c|ccccl|c|c}
\toprule
\rule{0pt}{9pt} \multirow{2}{*}{Method} & \multirow{2}{*}{Pub.} & \multicolumn{5}{c|}{\textbf{UW-VOS}} & DAVIS$_{17}$ & YT-VOS$_{19}$  \\ %\cline{3-10}
\rule{0pt}{9pt} &  & $\mathcal{J}\&\dot{\mathcal{F}}$ & $\dot{\mathcal{F}}$ & $\mathcal{J}$ & $\mathcal{F}$ & \multicolumn{1}{c|}{$\mathcal{J}\&\mathcal{F}$}      & $\mathcal{J}\&\mathcal{F}$ & $\mathcal{J}\&\mathcal{F}$     \\
\midrule
STCN \cite{STCN} & \pub{NIPS'21} & 60.5 & 62.6 & 58.4 & 65.0 & 61.7\textcolor{gray}{(-22.4)}        & 85.4 & 82.7   \\
AOT-R50 \cite{AOT} & \pub{NIPS'21} & 73.5 & 76.3 & 70.7 & 78.7 & 74.7\textcolor{gray}{(-9.8)}        & 84.9 & 84.1   \\
DeAOT-R50 \cite{DeAOT} & \pub{NIPS'22} & 74.4 & 77.4 & 71.5 & 79.9 & 75.7\textcolor{gray}{(-9.9)}       & 85.2 & 86.0  \\
RDE \cite{RDE} & \pub{CVPR'22} & 58.7 & 61.5 & 56.0 & 64.0 & 60.0\textcolor{gray}{(-23.1)}       & 84.2 & 81.9  \\
XMem \cite{XMem} & \pub{ECCV'22} & 70.1 & 72.5 & 67.7 & 74.8 & 71.3\textcolor{gray}{(-14.6)}       & 86.2 & 85.6  \\
DEVA \cite{DEVA} & \pub{ICCV'23} & 76.7 & 79.2 & 74.1 & 81.4 & 77.8\textcolor{gray}{(-8.3)}       & 86.8 & 85.4 \\
Cutie-B \cite{Cutie} & \pub{CVPR'24} & 75.6 & 78.0 & 73.3 & 80.0 & 76.6\textcolor{gray}{(-11.3)}       & 88.8 & 87.0  \\
JointFormer \cite{JointFormer} & \pub{PAMI'25} & 74.1 & 76.5 & 71.6 & 78.1 & 74.9\textcolor{gray}{(-13.7)}       & 89.7 & 87.5  \\
SAM2-B+ \cite{SAM2} & \pub{ICLR'25} & 85.2 & 87.2 & 83.1 & 88.7 & 85.9\textcolor{gray}{(-3.5)}       & 90.2 & 88.6  \\
\bottomrule
\end{tabular}
\end{adjustbox}
} % end local settings
\label{tab:vos-zero-shot}
\end{table*}

\textbf{Cross-domain Generalization.}
To evaluate the cross-domain generalization of existing VOS methods, we conduct zero-shot evaluation on UW-VOS using models pretrained exclusively on open-air datasets (\ie, DAVIS 2017~\cite{DAVIS17} and YouTube-VOS 2019~\cite{YouTube-VOS}), without any fine-tuning on underwater data. As shown in \cref{tab:vos-zero-shot}, all methods exhibit substantial performance degradation when transferred to the underwater domain, with an average $\mathcal{J}\&\mathcal{F}$ drop of 13 points across nine methods. The degradation varies significantly: traditional memory-based approaches (\eg, STCN~\cite{STCN} and RDE~\cite{RDE}) suffer the most severe drops of 22--23 points, methods with hierarchical propagation (\eg, AOT~\cite{AOT}, DeAOT~\cite{DeAOT}, DEVA~\cite{DEVA}) show moderate drops of 8--10 points, while foundation model-based SAM2-B+~\cite{SAM2} demonstrates superior cross-domain robustness with only a 3.5-point drop. This performance gap stems from the unique underwater challenges absent in open-air datasets, including pronounced color shift toward the blue-green spectrum (\cf \cref{fig:channel_intensity}), high prevalence of small targets (56.3\%), and frequent camouflaged objects and similar distractors (\cf \cref{fig:attribute_instance}), underscoring the necessity of UW-VOS as a dedicated underwater benchmark.

\begin{table*}[tb]
\caption{
In-domain benchmark results of VOS methods trained and evaluated on UW-VOS.
$\ast$ indicates that due to hardware constraints, a reduced training budget was employed; see the Appendices for details. Inference speed (FPS) and GPU memory usage (GiB) are measured on a single 3090 GPU. For SAM2, video frames are offloaded to CPU memory to balance inference speed and memory usage. \textbf{Bold} denotes the best results.
}
\centering
{\fontsize{10pt}{3.6mm}\selectfont
\setlength{\tabcolsep}{2mm}

\begin{adjustbox}{max width=\textwidth}
\begin{tabular}{l|c|ccccccc}
\toprule
Method & Pub. & FPS & Mem. & $\mathcal{J}\&\dot{\mathcal{F}}$ & $\mathcal{J}$ & $\dot{\mathcal{F}}$ & $\mathcal{F}$ & $\mathcal{J}\&\mathcal{F}$   \\
\midrule
STCN \cite{STCN} & \pub{NIPS'21} & \textbf{51.4} & 14.8    & 65.9 & 63.6 & 68.3 & 70.6 & 67.1  \\
AOT-R50 \cite{AOT} & \pub{NIPS'21} & 30.7 & 2.0    & 77.3 & 74.6 & 80.1 & 82.2 & 78.4  \\
DeAOT-R50 \cite{DeAOT} & \pub{NIPS'22} & 45.4 & 1.8    & 78.0 & 75.0 & 81.0 & 83.1 & 79.0  \\
RDE \cite{RDE} & \pub{CVPR'22} & 15.3 & 2.2    & 65.0 & 62.6 & 67.4 & 69.6 & 66.1  \\
XMem \cite{XMem} & \pub{ECCV'22} & 36.8 & 2.3    & 76.1 & 73.8 & 78.4 & 80.6 & 77.2  \\
DEVA \cite{DEVA} & \pub{ICCV'23} & 45.6 & 2.2    & 80.2 & 77.9 & 82.5 & 84.6 & 81.3  \\
Cutie-B \cite{Cutie} & \pub{CVPR'24} & 45.4 & \textbf{1.3}    & 81.9 & 79.7 & 84.1 & 86.1 & 82.9  \\
JointFormer$^\ast $ \cite{JointFormer} & \pub{PAMI'25} & 2.2 & 9.4    & 75.9 & 73.4 & 78.4 & 79.9 & 76.7  \\
SAM2-B+ \cite{SAM2} & \pub{ICLR'25} & 19.3 & 2.7    & 86.8 & 84.7 & 88.8 & 90.2 & 87.5  \\
\midrule
$\mathrm{ARA_{XMem}}$ & - & 36.7 & 2.3    & 76.4 & 74.1 & 78.8 & 80.8 & 77.5  \\
$\mathrm{ARA_{\text{SAM2-B+}}}$ & - & 19.3 & 2.7    & \textbf{87.6} & \textbf{85.5} & \textbf{89.6} & \textbf{91.1} & \textbf{88.3}  \\
\bottomrule
\end{tabular}
\end{adjustbox}
} % end local settings
\label{tab:vos-benchmark}
\end{table*}

\textbf{In-domain Benchmark.}
To assess the effectiveness of domain-specific training and ARA-based adaptation, we train all methods on the UW-VOS training set and evaluate them on the validation set. As shown in \cref{tab:vos-benchmark}, in-domain training improves the performance of representative VOS methods compared with zero-shot cross-domain evaluation (\cf \cref{tab:vos-zero-shot}), but the gains vary substantially across architectures. Among non-foundation-model baselines, Cutie-B~\cite{Cutie} and DEVA~\cite{DEVA} achieve competitive $\mathcal{J}\&\dot{\mathcal{F}}$ scores of 81.9 and 80.2, respectively, while memory-based methods such as XMem~\cite{XMem} remain below 77 despite in-domain training. SAM2-B+~\cite{SAM2} provides the strongest baseline with 86.8 $\mathcal{J}\&\dot{\mathcal{F}}$, showing the benefit of large-scale foundation-model pretraining.

Instantiating ARA on different backbones further improves the corresponding host models without changing their temporal memory mechanisms. $\mathrm{ARA_{XMem}}$ improves XMem from 76.1 to 76.4, matching the full fine-tuning result while updating only the inserted restoration modules and task-specific head. The improvement is more pronounced for the stronger SAM2 backbone: $\mathrm{ARA_{\text{SAM2-B+}}}$ reaches 87.6 $\mathcal{J}\&\dot{\mathcal{F}}$ and 88.3 $\mathcal{J}\&\mathcal{F}$, outperforming full SAM2 fine-tuning by 0.8 points in the primary metric while keeping the same inference speed and memory footprint. These results suggest that ARA is not merely a SAM2-specific naming variant, but a feature-space aquatic restoration operator that can be instantiated across different VOS backbones, with larger gains when the host representation is sufficiently expressive.

\begin{table}[tb]
\caption{
Comparison between ARA and several PEFT methods based on SAM2. The training time and memory usage are measured in hours and GiB, respectively.
}
\centering
{\fontsize{10pt}{3.6mm}\selectfont
\setlength{\tabcolsep}{2mm}

\begin{adjustbox}{max width=\textwidth}
\begin{tabular}{l|cc|cc|cc}
\toprule
\rule{0pt}{9pt} \multirow{2}{*}{Method}      & \multirow{2}{*}{\makecell{Total\\Params.}}    & \multirow{2}{*}{\makecell{Trainable\\Params.}}  & \multicolumn{2}{c|}{Training}     & \multirow{2}{*}{$\mathcal{J}\&\dot{\mathcal{F}}$}    & \multirow{2}{*}{$\Delta$}    \\ \cline{4-5}
\rule{0pt}{9pt}  &  &  & Time & Mem.    &     &     \\
\midrule
Full finetune & 80.8\,M & 80.8\,M & 5.35 & 14.56 & 86.8 & 0.0 \\
\midrule
BitFit \cite{BitFit} & 80.8\,M & 0.1\,M & 3.03 & 9.83 & 85.0 & $-1.8$ \\
LoRA \cite{LoRA} & 83.3\,M & 2.5\,M & 3.15 & 9.31 & 85.2 & $-1.6$ \\
AdaptFormer \cite{AdaptFormer} & 83.4\,M & 2.5\,M & 2.90 & 7.30 & 85.4 & $-1.4$ \\
\midrule
ARA & 82.4\,M & 1.6\,M & 3.05 & 6.83 & 87.6 & $+0.8$ \\
\bottomrule
\end{tabular}
\end{adjustbox}
} % end local settings
\label{tab:ablation}
\end{table}

\textbf{Comparison with PEFT Baselines.}
\Cref{tab:ablation} compares ARA with representative parameter-efficient fine-tuning strategies under the same SAM2-B+ backbone. Generic PEFT methods reduce the number of trainable parameters, but they underperform full fine-tuning in this underwater VOS setting: BitFit~\cite{BitFit}, LoRA~\cite{LoRA}, and AdaptFormer~\cite{AdaptFormer} obtain 85.0, 85.2, and 85.4 in $\mathcal{J}\&\dot{\mathcal{F}}$, respectively. In contrast, ARA reaches 87.6 with only 1.6\,M trainable parameters, exceeding full fine-tuning by 0.8 points while using roughly 2\% of the trainable parameters.

The efficiency profile also supports the motivation for a constrained adapter rather than unconstrained backbone updating. Compared with full fine-tuning, ARA reduces training time from 5.35 to 3.05 hours and GPU memory from 14.56 to 6.83\,GiB. Compared with LoRA and AdaptFormer, ARA uses fewer trainable parameters than both 2.5\,M-parameter alternatives and achieves a 2.2-point advantage over the strongest generic PEFT baseline. These results indicate that the gain does not come only from parameter efficiency; the feature-space inverse-restoration structure of ARA provides a more suitable inductive bias for aquatic domain shift than generic bias tuning, low-rank updates, or unconstrained bottleneck adapters.

\subsection{Attribute-Based Performance Analysis}

\begin{table*}[tb]
\caption{
Attribute-based performance analysis on UW-VOS validation set. We use $\mathcal{J}\&\dot{\mathcal{F}}$ as the evaluation metric. For each method, we highlight the \colorbox{blue!25}{lowest}, \colorbox{blue!15}{second lowest}, and \colorbox{blue!8}{third lowest} attribute scores.
}
\centering
{%\fontsize{8pt}{4mm}\selectfont
\setlength{\tabcolsep}{1mm}

\begin{adjustbox}{max width=\textwidth}
\begin{tabular}{l|ccccccccccccc}
\toprule
Method                          & Overall & ST      & FM    & SV    & VC    & OCC   & AC    & SD    & IC    & MB    & ER    & CAM   & ARV\\
\midrule
STCN \cite{STCN}                & 65.9    & \worst{37.0}    & 64.1  & 64.0  & 63.2  & 63.2  & 64.7  & 60.6  & 65.2  & \secondworst{55.3}  & \thirdworst{58.7}  & 59.1  & 64.5\\
AOT-R50 \cite{AOT}              & 77.3    & \worst{52.8}    & 77.2  & 76.2  & 75.2  & 73.9  & 76.8  & 74.6  & 77.6  & 73.1  & \thirdworst{65.4}  & \secondworst{60.1}  & 75.4\\
DeAOT-R50 \cite{DeAOT}          & 78.0    & \worst{58.5}    & 78.2  & 76.8  & 76.1  & 74.6  & 77.8  & 74.7  & 78.1  & 73.0  & \thirdworst{66.4}  & \secondworst{60.4}  & 76.4\\
RDE \cite{RDE}                  & 65.0    & \worst{43.4}    & 62.8  & 63.0  & 62.2  & 60.9  & 64.3  & 59.1  & 62.9  & 56.3  & \secondworst{55.1}  & \thirdworst{56.1}  & 63.4\\
XMem \cite{XMem}                & 76.1    & \worst{44.9}    & 74.7  & 74.7  & 73.0  & 73.5  & 74.4  & 72.8  & 74.5  & \thirdworst{66.9}  & 70.7  & \secondworst{62.6}  & 74.6\\
DEVA \cite{DEVA}                & 80.2    & \worst{56.6}    & 78.9  & 79.0  & 77.8  & 77.6  & 79.0  & 77.6  & 78.9  & 72.6  & \thirdworst{72.1}  & \secondworst{60.5}  & 78.9\\
Cutie-B \cite{Cutie}            & 81.9    & \worst{66.5}    & 79.4  & 80.9  & 79.7  & 79.7  & 80.9  & 79.5  & 80.6  & 76.7  & \thirdworst{76.0}  & \secondworst{70.5}  & 81.0\\
JointFormer \cite{JointFormer}  & 75.9    & \worst{53.1}    & 75.4  & 74.4  & 74.4  & 71.9  & 75.4  & 73.4  & 76.3  & 71.8  & \thirdworst{66.1}  & \secondworst{59.4}  & 74.6\\
SAM2-B+ \cite{SAM2}             & 86.8    & 85.1    & 85.6  & 86.1  & 84.7  & 84.4  & 86.0  & 84.7  & 86.6  & \thirdworst{83.4}  & \secondworst{77.4}  & \worst{74.0}  & 85.9\\
\bottomrule
\end{tabular}
\end{adjustbox}
} % end local settings
\label{tab:attribute}
\end{table*}

\begin{figure}[tb]
  \centering
  \includegraphics[width=0.86\textwidth]{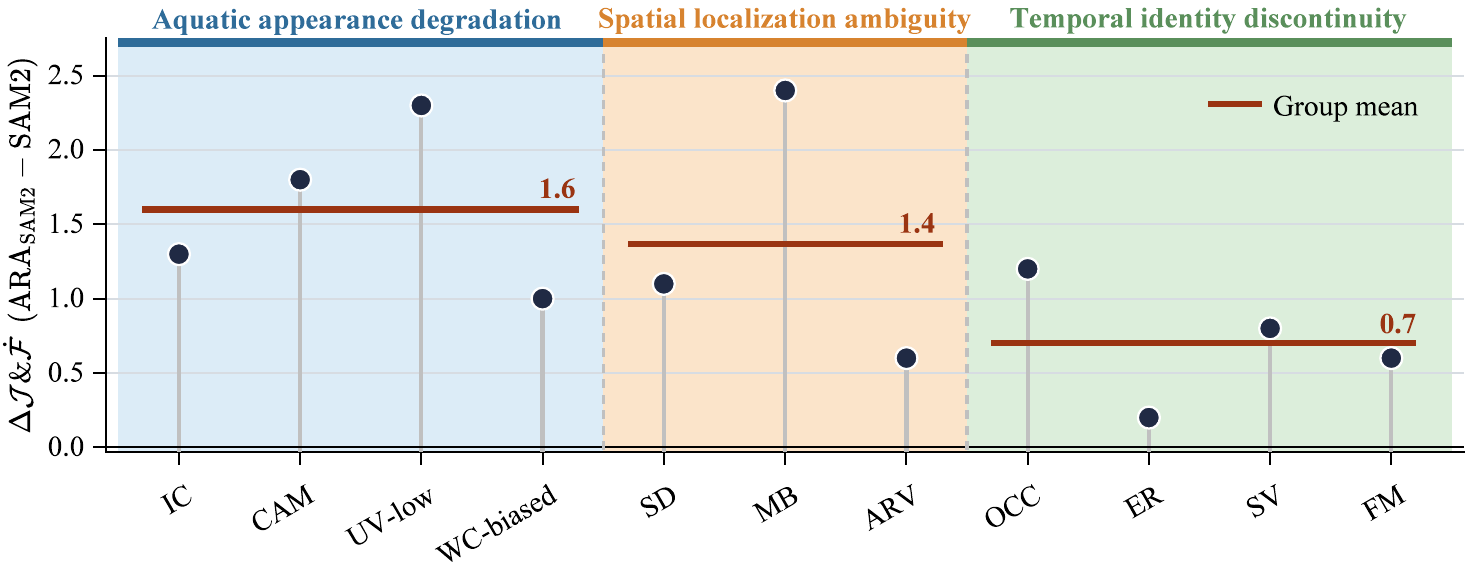}
  \caption{Attribute-level improvements of $\mathrm{ARA_{\text{SAM2-B+}}}$ over SAM2 on UW-VOS. $\Delta$ denotes the $\mathcal{J}\&\dot{\mathcal{F}}$ difference between $\mathrm{ARA_{\text{SAM2-B+}}}$ and SAM2 on each attribute subset.}
  \label{fig:attribute_group}
\end{figure}

To provide fine-grained insights into model behavior under diverse underwater conditions, we conduct attribute-based analysis on the UW-VOS validation set. \Cref{tab:attribute} reports the absolute performance of representative VOS methods across 12 binary attributes, while \cref{fig:attribute_group} groups the attribute-level gains of ARA over SAM2 according to the three failure factors discussed in \cref{sec:intro,fig:ara}. Specifically, aquatic appearance degradation captures impaired visibility and corrupted appearance cues; spatial localization ambiguity reflects weakened target--background separation and boundary evidence; and temporal identity discontinuity characterizes disrupted cross-frame association. These groups describe the dominant failure mechanism associated with each condition rather than mutually exclusive partitions, since one condition may affect multiple aspects of VOS. For the grouped analysis, we additionally split the multi-valued underwater visibility (UV) and water color (WC) annotations into low-visibility and biased-water-color subsets. This organization separates general method robustness from the specific effect of ARA-based feature-space restoration.

As shown in \cref{tab:attribute}, small target (ST) consistently emerges as the most challenging attribute for non-foundation-model methods, with performance drops of 19--29 points relative to overall scores. Memory-based approaches (\eg, STCN~\cite{STCN}, RDE~\cite{RDE}, and XMem~\cite{XMem}) further suffer from motion blur (MB) and exit-re-entry (ER), while hierarchical propagation methods (\eg, AOT~\cite{AOT}, DeAOT~\cite{DeAOT}, DEVA~\cite{DEVA}, and JointFormer~\cite{JointFormer}) frequently rank camouflage (CAM) and ER among their most difficult attributes. SAM2~\cite{SAM2} is substantially more robust on small targets, but still exhibits clear weakness on CAM and ER, indicating that underwater VOS failures are not explained by model capacity alone.

\Cref{fig:attribute_group} further analyzes whether ARA improves the attributes most related to its feature-space restoration design. For aquatic appearance degradation, ARA improves SAM2 by 1.3, 1.8, 2.3, and 1.0 points on IC, CAM, low underwater visibility, and biased water color, respectively, with an average gain of 1.6 points. For spatial localization ambiguity, ARA improves SD, MB, and ARV by 1.1, 2.4, and 0.6 points, respectively, averaging 1.4 points. These gains are consistent with the ARA formulation in \cref{eq:underwater-inverse,eq:ara-restore}: correcting veiling light and inverse-transmission effects improves the internal feature contrast used for appearance matching and local mask localization, without applying image-level enhancement.

In contrast, ARA brings smaller gains on temporal identity discontinuity, with improvements of 1.2, 0.2, 0.8, and 0.6 points on OCC, ER, SV, and FM, respectively, averaging 0.7 points. These limited gains reflect ARA's controlled scope in \cref{fig:ara}: it does not modify the host model's memory read/write or re-identification mechanisms, which are central to handling OCC and ER. Promising directions include occlusion-aware memory updates that suppress unreliable observations and long-term object-centric retrieval that combines appearance prototypes with motion cues to re-identify targets after re-entry.

\subsection{Transfer Learning}

\begin{wrapfigure}{r}{0.42\textwidth}
  \centering
  \vspace{-12pt}
  \includegraphics[width=0.4\textwidth]{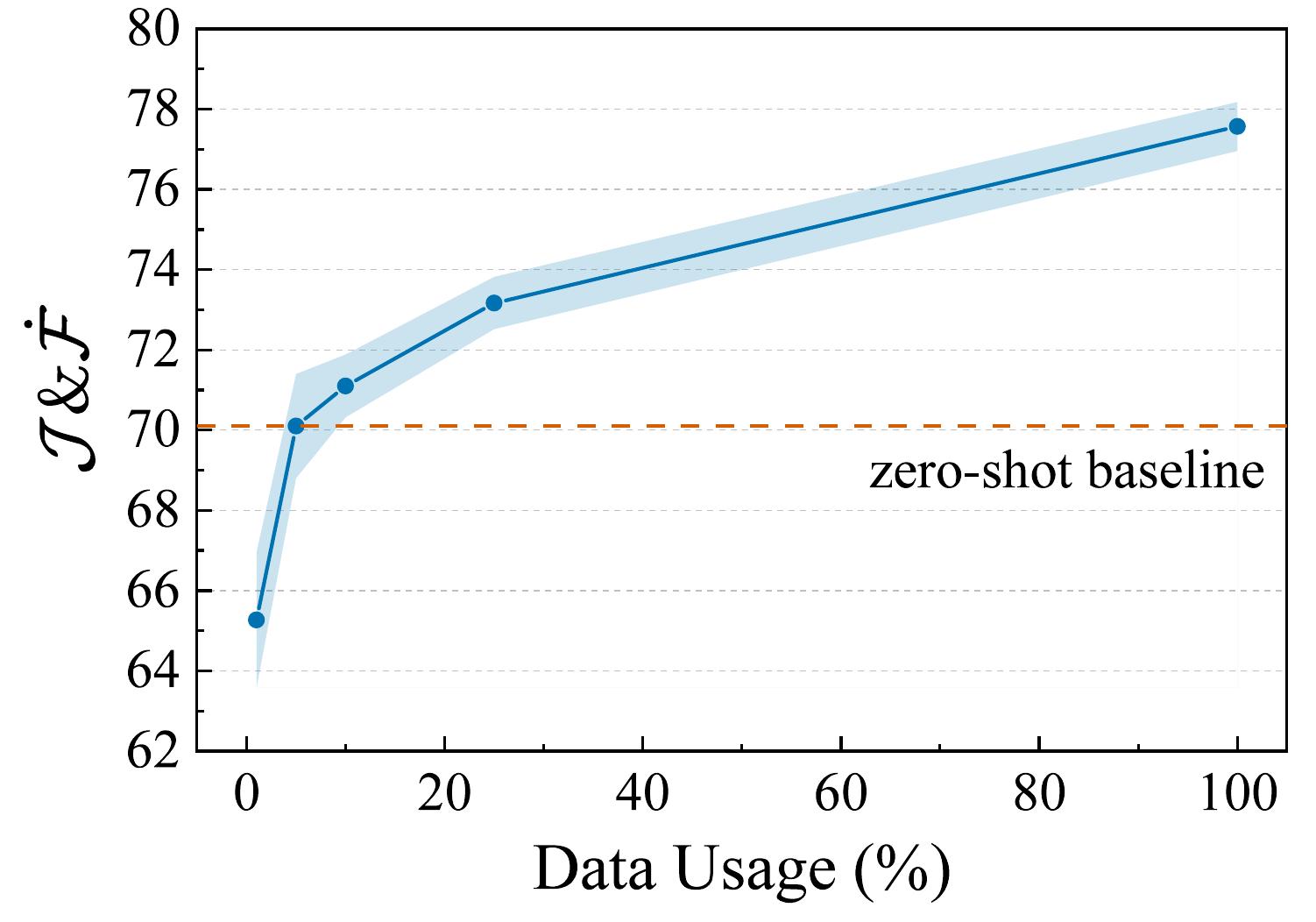}
  \caption{Cross-domain adaptation performance on UW-VOS with varying amounts of training data.}
  \label{fig:transfer_learning}
  \vspace{-12pt}
\end{wrapfigure}
To investigate annotation efficiency for domain adaptation, we fine-tune XMem~\cite{XMem} from open-air pretrained weights on varying subsets of the UW-VOS training set (1\%, 5\%, 10\%, 25\%, and 100\%). As shown in \cref{fig:transfer_learning}, fine-tuning with less than 5\% of data yields performance comparable to or even below the zero-shot baseline (70.1 in $\mathcal{J}\&\dot{\mathcal{F}}$). This pattern suggests possible negative transfer under highly limited supervision, where scarce domain-specific data may be insufficient to support effective adaptation. Performance improves as more training data are used, reaching 73.2 at 25\% and 77.4 with the full training set. Among the evaluated settings, a clear improvement over the zero-shot baseline emerges by the 25\% setting, while the 4.2-point gain from 25\% to 100\% indicates that comprehensive annotation remains beneficial. These results highlight UW-VOS's value not only as a benchmark but also as a resource for studying data efficiency in domain adaptation.

\subsection{Qualitative Analysis}

\begin{figure}[tb]
  \centering
  \includegraphics[width=\textwidth]{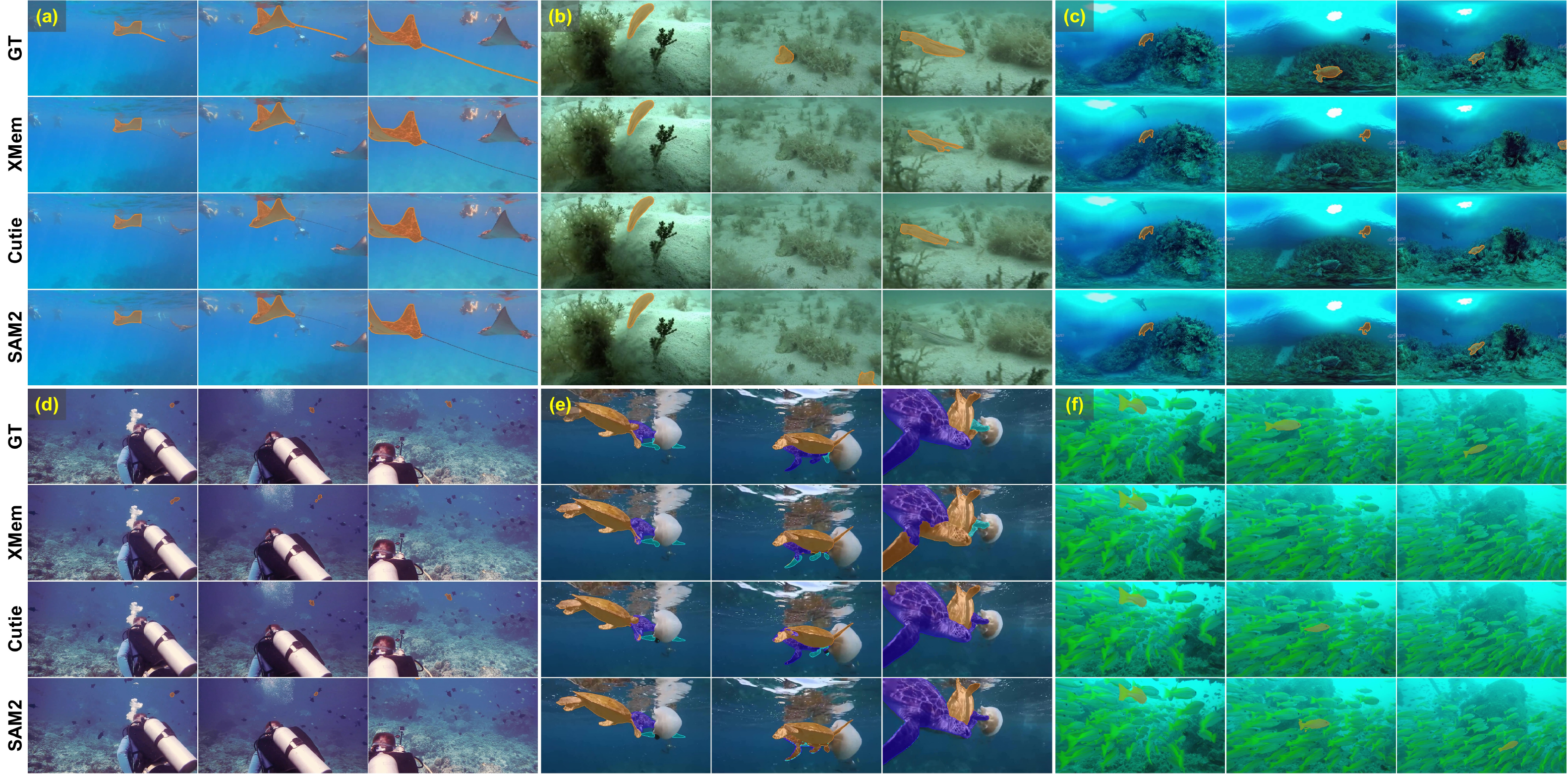}
  \caption{Qualitative results on UW-VOS. We compare the performance of XMem~\cite{XMem}, Cutie~\cite{Cutie}, and SAM2~\cite{SAM2} on six challenging cases. These cases include fine-grained structures (a), camouflage (b), target exit and re-entry (c), small objects (d), complex interactions (e), and crowded scenes (f).}
  \label{fig:qualitative_analysis}
\end{figure}

\Cref{fig:qualitative_analysis} presents six representative cases that illustrate the key limitations of existing VOS methods in underwater scenarios. (1) Existing methods fail to accurately delineate fine-grained structures such as elongated objects (case a). (2) Even foundation models trained on massive datasets (\ie, SAM2~\cite{SAM2}) struggle to segment underwater camouflaged objects (case b). (3) When a target exits and subsequently re-enters the scene, models are prone to identity switches (case c). (4) Small-scale objects and numerous visually similar instances remain difficult for all evaluated methods (cases d and f). (5) Complex interactions involving frequent occlusions among multiple objects cause models to confuse the boundaries of different targets (case e). These qualitative observations are consistent with the quantitative findings in \cref{tab:attribute}, confirming that camouflage, small targets, and frequent exit-re-entry remain the most critical challenges for underwater VOS.

\section{Conclusion}

This paper studies underwater VOS as temporal dense segmentation under aquatic domain shift. We introduced UW-VOS, a large-scale diagnostic benchmark with 1,431 videos, 411 categories, 307,267 mask annotations, and 16 fine-grained attributes, to support both standard VOS evaluation and factorized analysis of aquatic appearance degradation, spatial localization ambiguity, and temporal identity discontinuity. We further proposed ARA, a physics-derived parameter-efficient adapter that transfers the inverse structure of underwater image formation into bottleneck feature space and can be instantiated in different VOS backbones. Experiments show that representative open-air VOS methods suffer a substantial cross-domain drop on UW-VOS, while ARA reaches comparable or better performance than full fine-tuning on both Transformer- and memory-based VOS backbones. Visualization and attribute-based analyses further indicate that feature-space aquatic restoration mainly benefits appearance and localization challenges.

The current study also reveals a clear boundary of spatial feature adaptation. Since ARA deliberately leaves the temporal memory mechanisms of its host models unchanged, failures caused by occlusion, exit-re-entry, and long-term identity discontinuity remain only partially addressed. These findings highlight the diagnostic value of UW-VOS beyond aggregate ranking: in addition to adapting spatial representations to aquatic image formation, future underwater VOS methods will require stronger temporal identity modeling and potentially multi-modal cues from underwater sensing platforms.

\clearpage
\appendix
\counterwithin*{figure}{section}
\counterwithin*{table}{section}

\section{Data Provenance}
\Cref{tab:data-provenance} summarizes the source composition of UW-VOS. All retained videos are filtered, deduplicated, and re-annotated into a unified VOS format. The WebUOT-1M-derived videos provide underwater video content and tracking-style source annotations, but not pixel-level VOS masks in our target format; therefore, UW-VOS re-annotates these videos with dense masks and attributes. The newly mined Internet videos are collected from copyright-free sources and annotated from scratch. This strategy separates video acquisition from annotation creation: existing public videos provide diverse underwater content, while the costly and quality-critical pixel-level masks, attributes, and verification records are newly produced by UW-VOS.

\begin{table}[h]
  \centering
  \caption{Data provenance of UW-VOS. ``Source masks'' denotes pixel-level VOS masks available from the original source, while ``UW-VOS masks'' denotes the final verified masks released in our benchmark.}
  \label{tab:data-provenance}
  \setlength{\tabcolsep}{1.5mm}
  \begin{adjustbox}{max width=\textwidth}
    \begin{tabular}{lccccc}
      \toprule
      Source & Videos & Frames & Source masks & UW-VOS masks & Categories \\
      \midrule
      WebUOT-1M-derived & 1,032 & 223,996 & 0 & 223,996 & 408  \\
      Newly mined Internet videos & 399 & 63,203 & 0 & 83,271 & 179  \\
      \midrule
      Total & 1,431 & 287,199 & 0 & 307,267 & 411 \\
      \bottomrule
    \end{tabular}
  \end{adjustbox}
\end{table}

\section{Attribute Definitions}
\Cref{tab:attributes} provides the complete definitions of all attributes used in UW-VOS. We adopt a subset of attributes from WebUOT-1M~\cite{WEBUOT1M} with further adjustments and extensions to better characterize underwater scenarios. Each video sequence is annotated with multiple attributes to enable fine-grained analysis of method performance under specific challenges. The attributes are organized into two groups: generic attributes commonly used in video object segmentation, and underwater-specific attributes that capture domain-particular challenges.

\begin{table}[p]
	\centering
    \caption{Definitions of attributes in UW-VOS. We adopt a subset of attributes from WebUOT-1M~\cite{WEBUOT1M} with further adjustments and extensions. The top part presents generic attributes for video object segmentation, while the bottom part highlights several challenging attributes specific to underwater scenarios.}
    \label{tab:attributes}
    {
    \fontsize{10pt}{3.6mm}\selectfont
    \setlength{\tabcolsep}{4pt}

    \begin{adjustbox}{max width=\textwidth}
	  \begin{tabular}{l|p{0.92\textwidth}}
        \toprule
        Attr. & Definition  \\
        \midrule
        ST    & \textit{Small Target.} The average mask size across the sequence is below 0.1\% of the image area. \\
        \rowcolor{gray!20}
        FM    & \textit{Fast Motion.} The average displacement of the target center between consecutive frames exceeds 20 pixels. \\
        SV    & \textit{Scale Variation.} The target mask area ratio between any two frames falls outside the range {[}0.5,~2.0{]}. \\
        \rowcolor{gray!20}
        VC    & \textit{Viewpoint Change.} The object sequence contains multiple camera shots or viewpoint transitions. \\
        OCC   & \textit{Occlusion.} The target object is partially or fully occluded during the sequence. \\
        \rowcolor{gray!20}
        AC    & \textit{Appearance Change.} The target undergoes significant appearance variation due to rotation or deformation. \\
        SD    & \textit{Similar Distractors.} Visually similar objects appear in proximity to the target. \\
        \rowcolor{gray!20}
        IC    & \textit{Illumination Change.} The scene exhibits substantial illumination variation. \\
        MB    & \textit{Motion Blur.} The target region is blurred due to object motion or camera movement. \\
        \rowcolor{gray!20}
        MT    & \textit{Multiple Targets.} The video contains multiple objects requiring segmentation. \\
        ER    & \textit{Exit and Re-entry.} The target exits and subsequently re-enters the field of view. \\
        \midrule
        \rowcolor{gray!20}
        CAM    & \textit{Camouflage.} The target exhibits visual similarity to the surrounding background. \\
        ARV    & \textit{Aspect Ratio Variation.} The target bounding box aspect ratio falls outside the range {[}0.5,~2.0{]}. \\
        \rowcolor{gray!20}
        UV     & \textit{Underwater Visibility.} The visibility of the target region. (low, medium, high) \\
        US     & \textit{Underwater Scenes.} The specific underwater environment in the sequence. (sea, river, pool, water tank, fish tank, basin, bowl, cup, aquarium, pond, puddle, lake) \\
        \rowcolor{gray!20}
        WC     & \textit{Water Color.} The color characteristics of the water surrounding the target. (colorless, ash, gray, green, light green, dark, blue-black, deep blue, blue, light blue, partly blue, gray-blue, light yellow, light brown, cyan, light purple) \\
        \bottomrule
		\end{tabular}
    \end{adjustbox}
    }
\end{table}

\clearpage
\section{Experiment Details}
\subsection{Implementation Details of JointFormer}
Due to GPU memory limitations, when training on UW-VOS we use a reduced training budget compared with the default setting. Specifically, we reduce the effective batch size from 8 to 4, shorten the training clip from 4 frames (2 reference frames) to 3 frames (2 reference frames). Other training settings remain unchanged. During inference, we also use the default setting.

\subsection{Implementation Details of ARA}
For the SAM2 instantiation of ARA, we build on the SAM2-B+ framework with the Hiera-B+ image encoder and initialize the model from the official pretrained weights. ARA is inserted into blocks of Stage~3 and Stage~4, where domain-sensitive semantic and object-level features are more strongly encoded. Each selected block contains an ARA branch on the attention residual stream and another ARA branch on the feed-forward stream. During fine-tuning, only the ARA parameters are updated, while the prompt encoder, mask decoder, temporal memory modules, and the remaining backbone parameters are kept frozen. The input resolution is set to $1024 \times 1024$.

For the XMem instantiation, ARA is attached to the spatial feature stream before memory key-value construction. This placement allows the features written to and retrieved from memory to be adapted to underwater appearance degradation, while keeping the memory read, memory update, and temporal propagation modules unchanged. We follow the same UW-VOS training protocol as the XMem in-domain baseline and update only the inserted ARA parameters in addition to the task-specific segmentation head required by the original framework. This setting isolates the effect of ARA-based feature-space restoration from changes in temporal memory design.

For the SAM2 instantiation, the model is fine-tuned on the UW-VOS training set for 40 epochs. We employ the AdamW optimizer with a base learning rate of $5.0 \times 10^{-5}$. A cosine annealing schedule is used for learning rate decay, and weight decay is set to $0.01$. The total batch size is 4 distributed across 4 NVIDIA 3090 GPUs, with gradient clipping set to a maximum norm of $1.0$. We utilize mixed precision training  for computational efficiency. We sample video clips containing $T=4$ frames during training. Data augmentation includes random horizontal flips, random affine transformations, random resizing, and color jittering.

We optimize the model using a linear combination of Focal loss, Dice loss, and IoU loss with weights $\lambda_{\text{focal}}=20$, $\lambda_{\text{dice}}=1$, and $\lambda_{\text{iou}}=1$, respectively.

\section{More Results}

\begin{figure}[tb]
  \centering
  \includegraphics[width=0.7\textwidth]{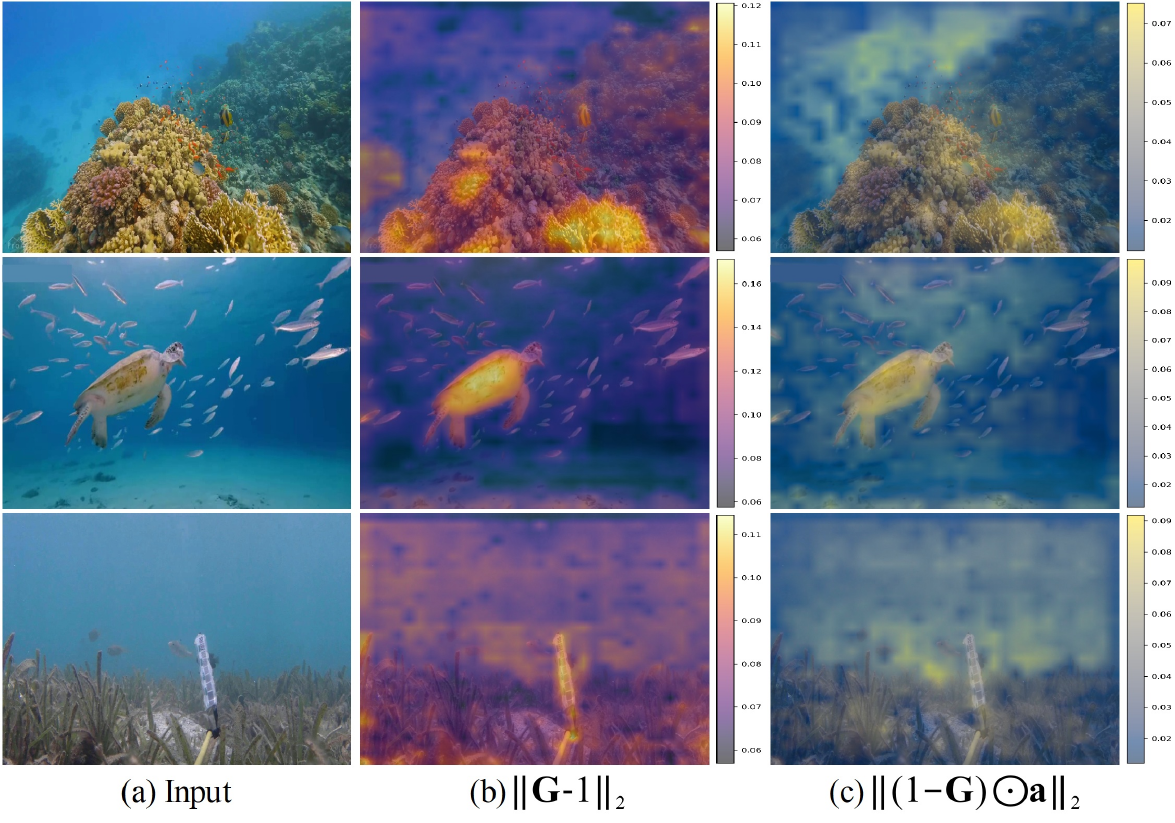}
  \setlength{\abovecaptionskip}{4pt}
  \caption{Visualization of ARA responses. Each row shows (a) the input image, (b) the channel-wise $\ell_2$ magnitude of the gain deviation from identity, $\mathbf{G}-1$, and (c) that of the effective additive contribution $(1-\mathbf{G})\odot\mathbf{a}$. Although $\mathbf{a}$ is shared across spatial locations, its effective contribution varies through $\mathbf{G}$. The gain-deviation response is concentrated around target regions, whereas the effective additive contribution is stronger in the surrounding water column and low-contrast background regions.}
  \label{fig:heatmap}
\end{figure}

\textbf{ARA Response Visualization.}
\Cref{fig:heatmap} visualizes the gain deviation $\mathbf{G}-1$ from identity and the effective additive contribution $(1-\mathbf{G})\odot\mathbf{a}$ in the expanded restoration form $\widehat{\mathbf{Z}}=\mathbf{G}\odot\mathbf{Z}+(1-\mathbf{G})\odot\mathbf{a}$. The gain-deviation response is concentrated around target regions and discriminative object parts, whereas the effective additive contribution varies spatially through $\mathbf{G}$ despite the shared $\mathbf{a}$ and is stronger in the surrounding water column and low-contrast background regions. These complementary patterns are consistent with ARA's constrained feature-space restoration: $\mathbf{G}-1$ indicates content-dependent departures from identity modulation, while $\mathbf{a}$ supplies a shared channel reference.

\clearpage
\section*{Acknowledgements}
This work is supported by the National Natural Science Foundation of China under Nos.~62276061 and 62436002. This work is also supported by Research Fund for Advanced Ocean Institute of Southeast University (Major Program MP202404).

\bibliographystyle{elsarticle-num}
\bibliography{main}

\end{document}